# Automatic Road Subsurface Distress Recognition from Ground Penetrating Radar Images using Deep Learning-based Cross-verification

**Chang Peng [a], Bao Yang [a], Meiqi Li [b], Ge Zhang [b, c, *], Hui Sun [b, *], Zhenyu Jiang [a, *]**

[a] Department of Engineering Mechanics, School of Civil Engineering and Transportation, South China University of Technology, Guangzhou 510640, China

[b] Guangdong Provincial Academy of Building Research Group Co. Ltd., Guangzhou 510510, China

[c] School of Civil and Transportation Engineering, Guangdong University of Technology, Guangzhou 510006

* Corresponding author. E-mail addresses: gezhang@gdut.edu.cn (G. Zhang), 26564923@qq.com (H. Sun), zhenyujiang@scut.edu.cn (Z. Jiang).

## Abstract

Ground penetrating radar (GPR) has become a rapid and non-destructive solution for road subsurface distress (RSD) detection. However, recognizing RSD from GPR images is labor-intensive and heavily relies on the expertise of inspectors. Deep learning-based automatic RSD recognition, though ameliorating the burden of data processing, suffers from insufficient capability to recognize defects. In this study, a novel cross-verification strategy was proposed to fully exploit the complementary abilities of region proposal networks in object recognition from different views of GPR images. Following this strategy, three YOLO-based models were used to detect the RSD (voids and loose structures) and manholes. Each model was trained with a specific view of 3D GPR dataset, which contains rigorously validated 2134 samples of diverse types obtained through field scanning. The cross-verification strategy achieves outstanding accuracy with a recall of over 98.6% in the tests using real field-scanning data. Field tests also show that deep learning-based automatic RSD recognition can reduce the human labor of inspection by around 90%.

**Keywords:** Ground penetrating radar; Road subsurface distress; Cross-verification; Deep learning; YOLO model; Field scanning

## 1. Introduction

Road subsurface distress (RSD) is the structural failures occurring in multiple zones beneath the pavement (such as underlayer, subgrade, and soil foundation) [1]. Typical RSD manifests itself as loose structures, interlayer debonding and voids. These defects, once formed, are difficult to repair through routine maintenance methods. The consequences include recurrent deterioration, substantial reduction in load bearing capacity of road, and risk of catastrophic road collapse [2]. In recent years, RSD-induced road collapse incidents have been increasingly reported worldwide [3,4], particularly in highly urbanized regions with intensive traffic networks. These accidents bring not only enormous damage to transportation infrastructure but also grave threats to public safety. Timely detection of RSD has therefore become a pressing issue in road engineering.

Ground penetrating radar (GPR), a representative of non-destructive technologies to survey underground structures, has gained widespread applications in the health monitoring of infrastructure (including roads, bridges, and tunnels), due to its high detection accuracy and efficiency, as well as low cost [5]. GPR emits electromagnetic waves into the ground, which can be reflected by the interfaces between the materials with different dielectric constants (Fig. 1a). The underground structure can be reconstructed through analyzing the return signals [6]. The original data acquired by GPR is A-scan (one-dimensional array), which represents the signal amplitude variation with depth for each radar channel. Multiple A-scan channels collected along the scanning distance of a mobile GPR surveying system form a 3D matrix (illustrated in Fig. 1b), with slices in three orthogonal planes as B-scan (longitudinal section), C-scan (horizontal section), and D-scan (transverse section).

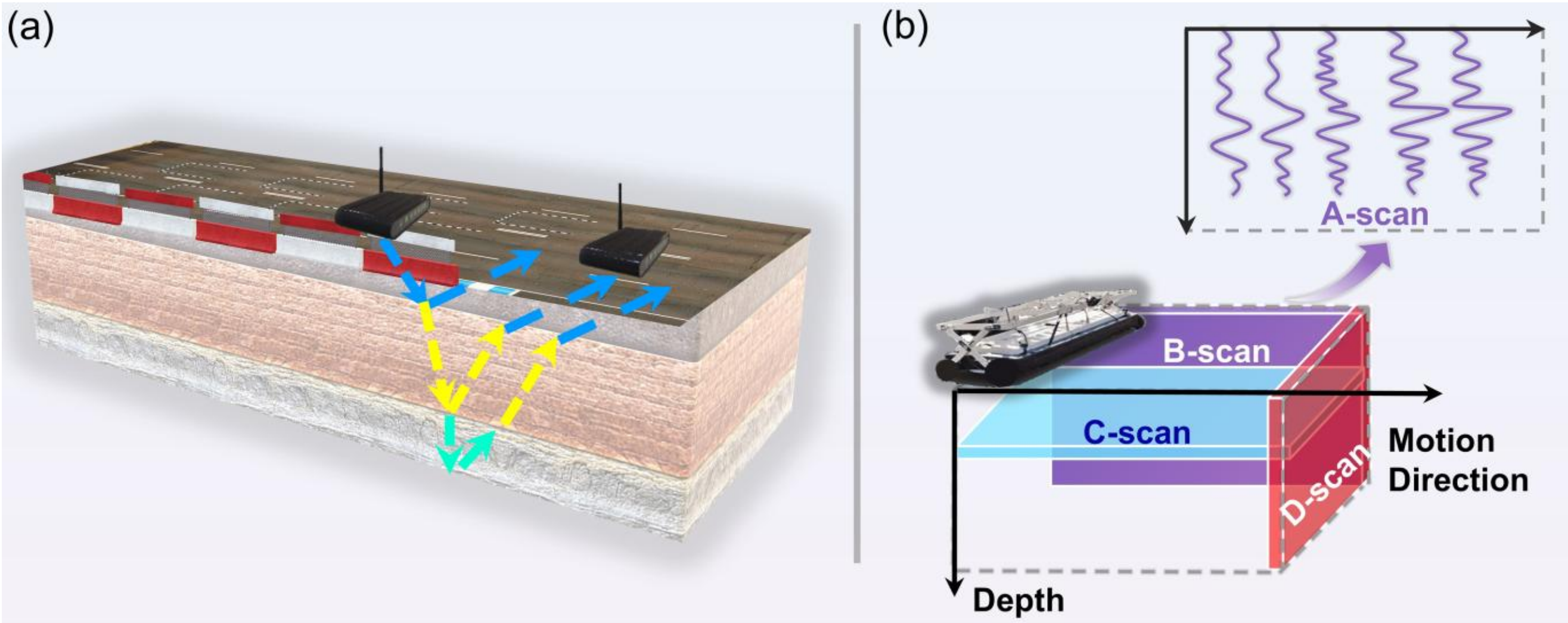

**Figure 1.** Schematic diagram of GPR. (a) Illustration of radar signal transmission and reception of electromagnetic waves in a multi-layered road structure. (b) Schematic diagram of four scans of a 3D matrix of GPR data.

The images reconstructed from GPR signals, unlike the ones collected through optical imaging techniques, are hard to analyze or interpret [7]. Identification of anomalous signals relies heavily on the expertise of inspectors, which makes manual processing time-consuming and labor-intensive, while also limits the objectivity of this task [8,9]. It was estimated that manual processing of GPR images for one kilometer of road takes approximately one hour for professionals [7,10]. This is far from the current demand for large-scale and long-term RSD inspection.

Existing study on automatic interpretation of GPR signals can be categorized into two approaches: Early rule-based recognition was developed based on electromagnetic wave theory and heuristic algorithms to identify abnormal patterns in signals. Hough transform filters [11] and hyperbolic curve regression [12] were employed to extract features in frequency domain or time-frequency domain for further analysis and comparison [13]. Rule-based methods were found to work well in ideal experimental conditions, but their effectiveness diminishes dramatically in practical applications where electromagnetic signals are distorted by various controlling factors.

Recent data-driven recognition benefited from the advancement of artificial intelligence and its integration into engineering. In comparison with traditional methods, machine learning-based approaches, such as support vector machines [14], K-nearest neighbors [15], hidden Markov model [16] and naive

Bayes model [17], were found adept at establishing correlation between RSD and ambiguous features in GPR signals. Machine learning models can extract multidimensional features to improve detection accuracy and applicability. Deep learning-based approaches, with greatly enhanced ability to extract complex features from waveforms and images, achieve significantly improved performance in RSD recognition [3]. Encouraged by the success of deep learning in image recognition, various object identification models have been applied to the analysis of B-scan image. AlexNet-based architectures were first applied to the classification of B-scan images [18], laying the foundation for subsequent research. More sophisticated networks like Bi-LSTM and residual CNN showed better performance [19]. Recently, attention mechanism and transfer learning were introduced to increase the recognition accuracy through guiding the models focus on key regions [20].

Object detection models, combining the functions of recognition and segmentation, meet well the requirement of GPR image analysis in engineering applications. Faster region-based convolutional neural network (R-CNN), a representative two-stage detector, demonstrated considerable improvement in detection accuracy through introducing region proposal networks [21]. Single-stage detectors, represented by YOLO [22], became a popular research paradigm due to its higher efficiency [23–29]. Although the latest Transformer-based hierarchical medium inversion network [30] outperformed YOLO-based models when handling overlapped targets, the huge computational cost limit its practical application.

Three-dimensional radar data provides more comprehensive information on subsurface structures. 3D convolutional neural networks (3D CNN) were introduced recently for 3D GPR signal analysis [31–35]. 3D CNN-based approaches can capture volumetric features and multi-view correlations. However, they encounter several key limitations in practice: (i) Annotating 3D GPR field data is an arduous task. Most GPR software do not support the outputs in volumetric form, making it difficult for annotators to generate and verify accurate labels; (ii) The features in 3D GPR images are often highly complex, making it difficult for engineers to interpret and validate the results; (iii) It remains challenging to integrate these models with existing commercial software platforms, thereby limiting their deployment in current RSD detection systems.

To circumvent the difficulties of directly processing 3D GPR images, researchers attempted to stack the boundary boxes of abnormality in adjacent B-scan sections to form a cubic annotation region [31–35]. Another means is to conjoin the images of three scans (B/C/D scans) crossing at a specific location to a single image and feed it as input into neural networks to capture multi-view features [36–39]. The strategy was expected to overcome the inherent limitation of single-view methods and to enhance the accuracy of detection by exploiting the consistency among multiple views.

Existing multi-view data-driven methods for RSD detection in practical applications remains far from satisfactory. Table 1 lists the recent studies using field-scanning datasets as well as the performance achieved in RSD recognition. The identification accuracy (especially recall) of anomalous object reached by most of the reported work is below 90%, which cannot meet the need for practical road safety inspection. High accuracy was achieved only on a small and single-view dataset [32] or in the recognition of man-made objects (pipes and manholes) with significant and regular features [31,33,34]. In these studies, the datasets were relatively small (with tens or hundreds of samples) [35] or contain the samples with relatively poor diversity in object type or spot [31,35]. A large dataset with 9045 objects was constructed according to the field survey of highway in Jiangxi, China [40]. Unfortunately, the identification accuracy on this dataset is poor (with recall of 58.7%), indicating nonproficient or improper use of deep learning model. A very recent study built a field-scanning dataset comprising 1261 samples [41]. Nevertheless, the dataset does not fully capture the multi-view characteristics of distress as its annotations are restricted to B-scan.

Thus, the limited performance achieved by current studies could be ascribed to two factors: (i) Inadequate quantity and diversity of samples in training dataset hinder the full exploitation of model performance. (ii) Insufficient capability of network to distinguish RSD. This factor can also be considered as insufficient usage of information in multi-view images. Current multi-view analysis stays at the stage of simple concatenation of image features, neglecting the inherent characteristics in specific view and inter-view differences. Consequently, the increased dimensions of input data fail to effectively enhance feature extraction, whereas introduces potentially misleading information into the networks.

**Table1.** Summary of recent studies in deep learning-based RSD recognition.

| Model | Year | Data collection scenario | Object types | Sample numbers | Performance |
|---|---|---|---|---|---|
| 3D CNN [32] | 2020 | Urban roads in Seoul, Republic of Korea | Void, manhole, pipe, and subsoil | 64 from field survey, 2112 from data augment | Cavity (P=100%, R=87.5%)<br>Pipe (P=96.0%, R=100%)<br>Manhole (P=100%, R=100%)<br>Subsoil background (P=92.0%, R=100%) |
| 3D CNN [33] | 2020 | Roads in Nagano, Japan, with a total length of 230 km | Pipes with different directions | 3371 pipes | No pipe (P=84.6%, R=87.0%)<br>Trans. pipe (P=92.7%, R=94.1%)<br>Long. pipe (P=95.6%, R=88.6%) |
| 3D CNN [34] | 2022 | Roads in Japan, with a total length of 13 km | Void | 88 | P=91.6% , R=81.4% |
| 3D CNN [35] | 2022 | Airstrips of three Chinese international airports, with a total area of 21,083 m² | Void, crack, subsidence, and pipe | 6199 | Void (P=92%, R=89%)<br>Crack (P=73%, R=66%)<br>Subsidence (P=94%, R=87%)<br>Pipe (P=100%, R=100%) |
| YOLOX [40] | 2022 | Zhangshu-Ji'an Highway in Jiangxi, China, with a total length of 209.6 km | Pavement distress (not specified) | 9045 | P=87.71%, R=58.73% |
| 3D CNN [31] | 2023 | 14 roads in Beijing, Zhengzhou and Xining, China, with a total length of 213.504 km | Void, manhole, pipeline, and normal soils | 677 | A = 98.54% |
| MCGA-Net [41] | 2025 | Urban roads in Harbin, China, with a total length of 78.5 km | Cavity, Concave, Crack | 1261 from field survey, 2188 from data augment | P=92.8%, R=92.5% |

* A = Accuracy; P = Precision; R = Recall.

In this paper, a multi-view cross-verification strategy was proposed to improve the accuracy and reliability of region proposal networks in automatically recognizing RSD from GPR images. Three YOLO-based models were trained with B-scan, C-scan, and D-scan images separately from a large field-scanning dataset with annotations made by professionals. Each model was found to demonstrate high sensitivity to specific types of subsurface objects. The cross-verification strategy well exploits the information from the three views and demonstrates outstanding performance in field tests, satisfying the need for real road subsurface health monitoring. The rest of this paper is organized as follows: Section 2 elucidates our method, including the YOLO-based model for object detection, the cross-verification strategy, and the training dataset. Section 3 describes the procedure of network training. Section 4 evaluates the performance of our method using the dataset and new data collected from field tests. The paper is concluded in Section 5.

## 2. Methodology

### 2.1. Object detection model

The objects in the three views of GPR images are detected using a region proposal network based on a derivative of YOLO model [42], which is an anchor-free derivative achieving significant enhancements in both speed and accuracy. Fig. 2 illustrates the model architecture consisting of Backbone, Neck, and Head. Features are extracted by module Backbone composed of fundamental units including Focus [43], CBS (consisting of convolution, batch normalization and SILU activation function) [44], cross-stage partial network (CSP) [45], and spatial pyramid pooling (SPP) [44]. In CSP, Bottleneck (a structural unit in ResNet) [46] is utilized to enhance the non-linearity of the network architecture while reducing computational complexity. Three effective feature maps are generated in Backbone, each is a collection of numerous features at different scales, viz 1/8, 1/16, and 1/32 of the input image size.

In module Neck, the effective features are fused with upsampled and downsampled features to create three enhanced feature maps containing richer information. Module Head serves as the classifier and regressor to determine whether the features indicate specific objects. Three decoupled heads are employed to process the feature maps at three scales, respectively. Each generates three predictions: (i) Reg for determination of regression parameters to predict the bounding boxes; (ii) Obj for assessment of whether each feature point belongs to a specific object; (iii) Cls for identification of the object class. The three predictions are combined to generate output. Finally, the non-maximum suppression algorithm [47] is employed to eliminate redundant detections, identify optimal matches, and obtain the final detection results.

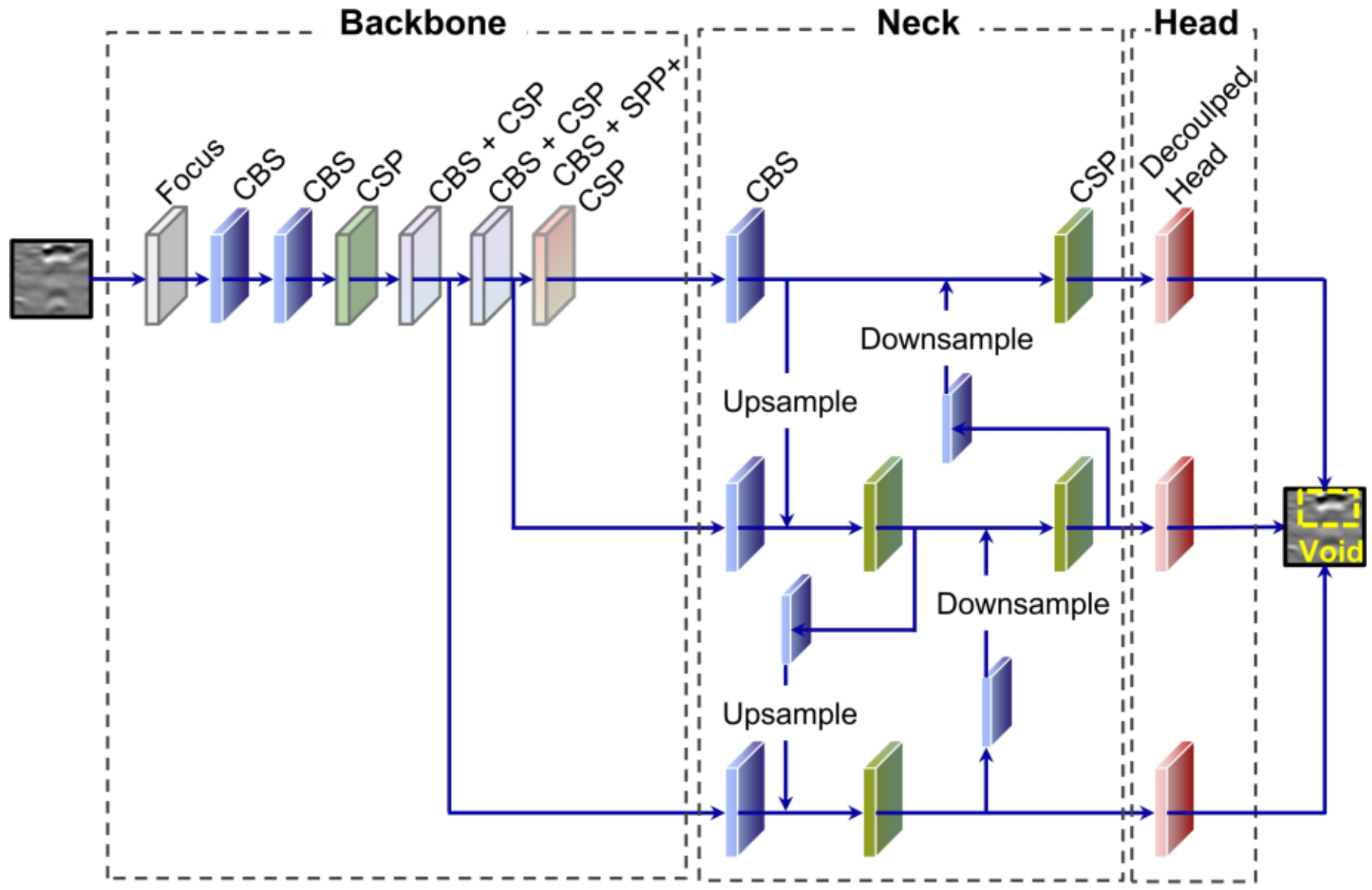

**Figure 2.** Architecture of YOLO-based model

### 2.2. Cross-verification strategy

Fig. 3 shows a typical sample of healthy part in C-scan, B-scan and D-scan, as well as the samples containing loose structure, void and manhole. Loose structure manifests itself in colormap of C-scan as fragmented speckles, whereas being indistinguishable in B-scan and D-scan. Void looks like a patch of irregular shape with sharp edge in C-scan. In the other two views, it appears as bell-shaped ripples. The ripples become pronounced in the case of void with large size. Manhole can be observed in C-scan as a patch of regular shape. In B-scan and D-scan, the corresponding ripples are similar to void but with stronger appearance and regular shape. It is noteworthy that cavities stemming from interlayer debonding are nearly identical to voids in GPR images. They are often treated as the same RSD in practical road safety inspections despite different forming mechanisms.

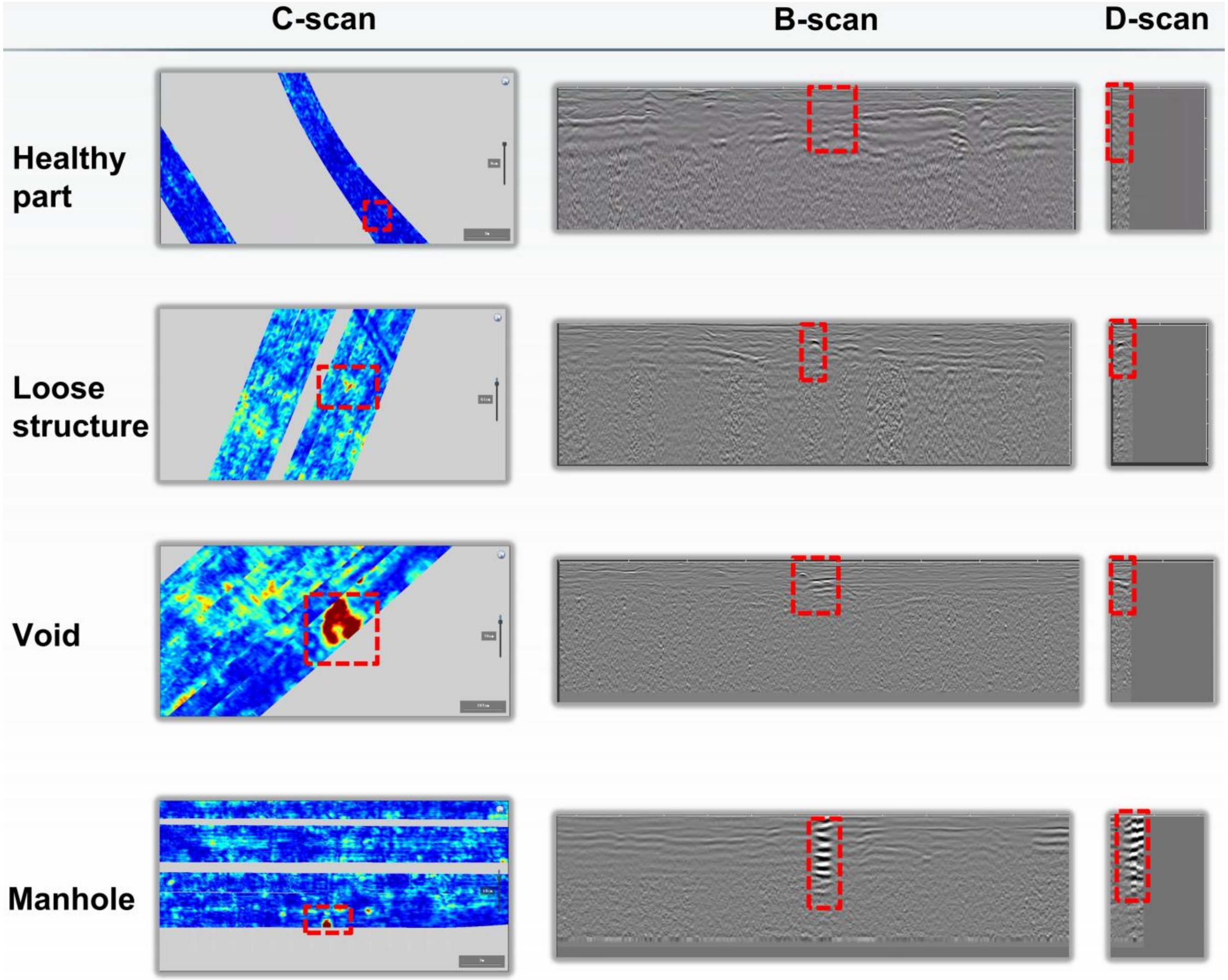

**Figure 3.** C-scan, B-scan and D-scan of typical road sections (healthy part, loose structure, void, and manhole).

It can be observed that anomalous structures are more distinguishable in C-scan (see examples in Fig. 3). The model trained with C-scan data (Model-C) demonstrates higher sensitivity to the difference between the spots with and without RSD (or manhole). Thus, the recognition of Model-C can serve as the criterion to determine the presence of RSD in view field. Recognition of Model-B (trained with B-scan data) is used to distinguish RSD from manholes, as the patterns in this section provide reliable reference. Recognition of Model-D (trained with D-scan data) is a necessary supplement to help differentiate between void and loose structure. Fig. 4 illustrates the workflow of the cross-verification strategy based on the recognition of multi-view scans. The strategy can be described as a three-step procedure: (i) Sifting out the healthy parts according to the recognition of Model-C; (ii) Filtering out manholes from the abnormal instances according to the recognition of Model-B; (iii) The remained defects are classified as voids or loose structures according to the recognition of Model-D.

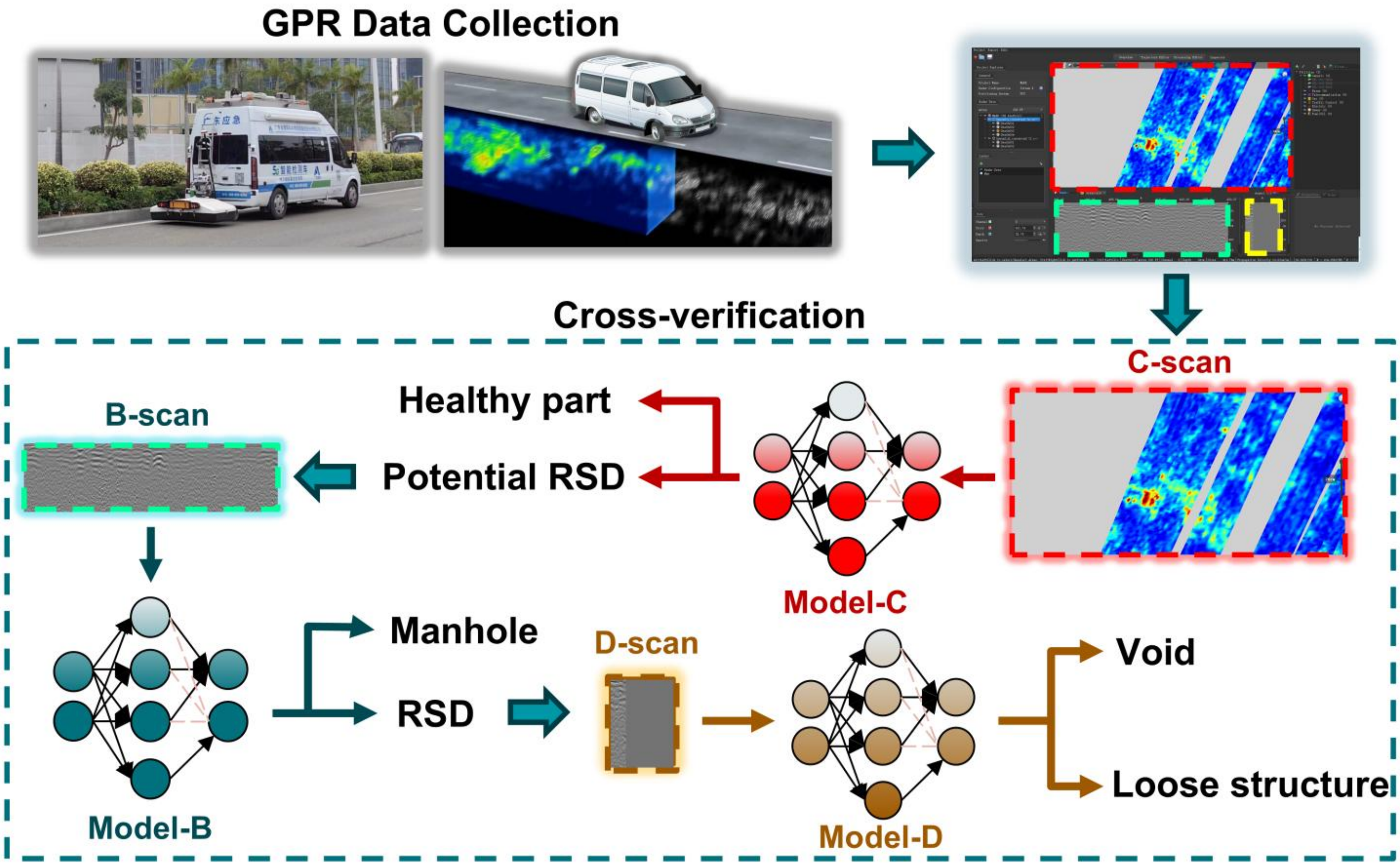

**Figure 4.** Flowchart of cross-verification strategy based on recognition from three views of GPR images.

It is noteworthy that the proposed cross-verification of three views may not guarantee 100% RSD recognition. Manual check is required to be conducted on the results of automatic processing. Thus, high recall, i.e. less false negative, is one of the most important targets for automatic recognition.

### 2.3. Construction of multi-view scan dataset

A large-scale dataset was constructed through field survey of urban roads in two metropolises of China: Chengdu and Guangzhou (as shown in Fig. 5). Both cities have populations exceeding twenty million residents and highly developed road transportation. GPR data were collected from 105 typical urban road sections (a total length of 1,250 kilometers) using utility vehicles equipped with StreamUP multi-channel GPR system. Table 2 lists the key parameters of GPR system and image acquisition. The sections were meticulously selected to cover a wide variety of road types, traffic volume, climatic conditions, and service age.

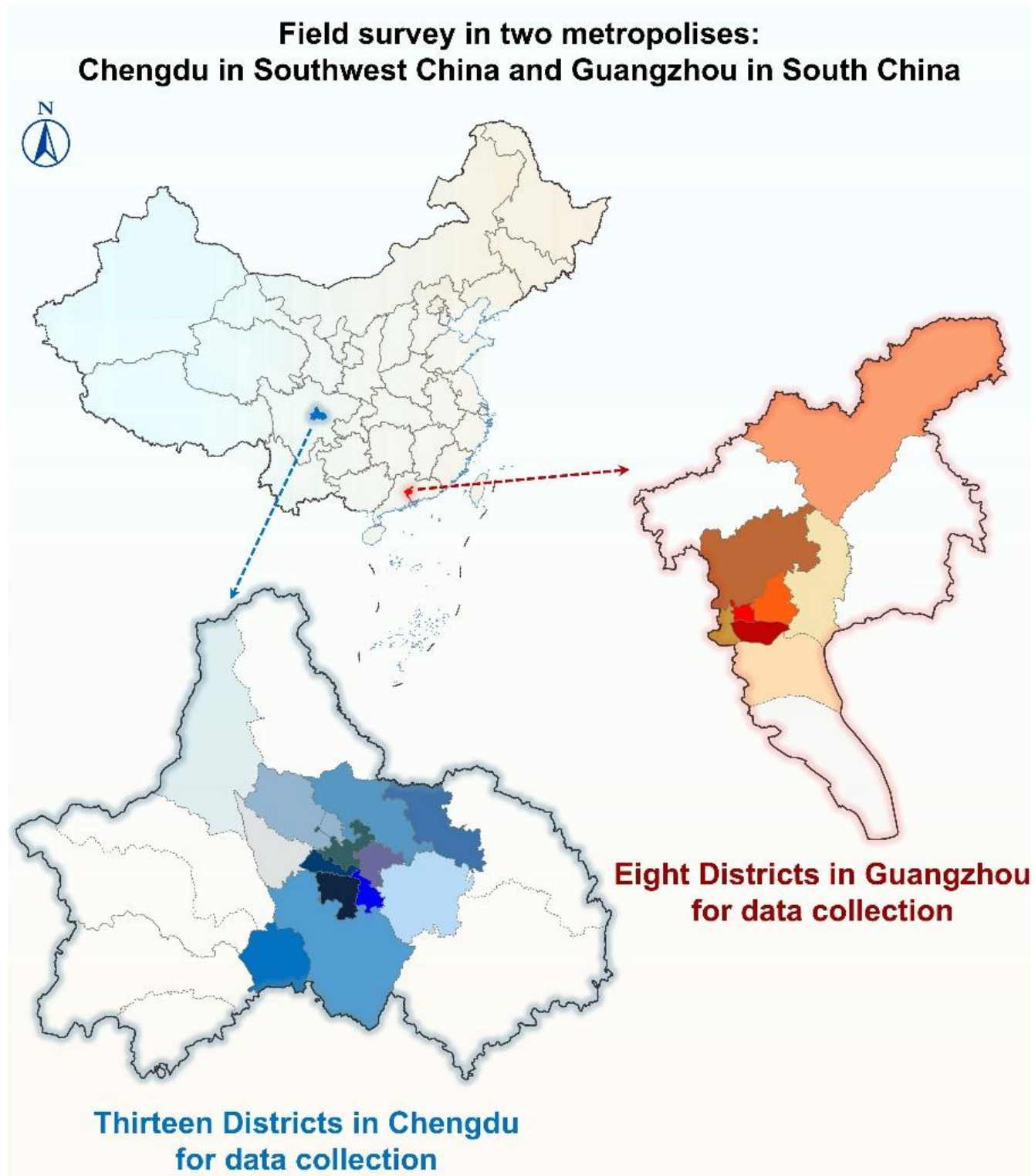

**Figure 5.** Regions for field dataset collection in two Chinese metropolises (Chengdu and Guangzhou).

**Table 2.** Key parameters of GPR system and image acquisition for construction of dataset.

| System parameter | Value | Acquisition parameter | Value |
|---|---|---|---|
| Maximum detection depth | 5 m | Minimum frequency | 200 MHz |
| Transverse sampling interval | 1.7 m | Maximum frequency | 600 MHz |
| Max number of samples | 512 | Maximum time-range | 180 ns |
| Maximum radar speed | ~150 km/h | | |

Table 3 shows that the field data covering four distinct landforms (alluvial plains, fluvial terraces, hilly terrain, and mountainous areas) and five major soil types (water-rich sandy pebble soil, expansive clay, silty clay, soft soil, and lateritic red soil). These roads exhibit substantial variation in geotechnical and soil properties, including groundwater content, compressibility, shear strength, settlement behavior, and susceptibility to deformation or collapse. The wide variety of road and subsurface conditions ensures that the dataset is sufficient for training and evaluating the models. The dataset includes 553 instances of healthy parts, 539 cases of void, 536 cases of loose structure, and 506 cases of manhole. All samples were meticulously annotated by experienced engineers, and most were verified through core sampling and falling-weight deflectometer test to ensure high reliability of labeling. Although minor uncertainties in annotation may be inevitable, their impact on the overall model performance could be quite limited.

**Table 3.** Geotechnical and soil characteristics of roads collected in the dataset.

| City | Landform | Representative soil | Soil properties | District |
|---|---|---|---|---|
| Chengdu | Alluvial plain | Water-rich sandy pebble soil | High groundwater content and loose structure; prone to void formation and sand collapse; unstable earth pressure with high compressibility and deformation under load | Jinniu; Chenghua; Jinjiang; Xindu; Wuhou; Qingyang; Wenjiang; Pidu; Xinjin |
| | Fluvial terrace | Expansive clay | Highly sensitive to moisture variation; prone to structural deformation | Shuangliu; Longquanyi; Qingbaijiang |
| | Mountainous area | Water-rich sandy pebble soil; silty clay | Highly compressible; poor permeability; prone to uneven settlement | Dujiangyan |
| Guangzhou | Alluvial plain | Soft soil | High water content; highly compressible; low shear strength; prone to uneven settlement and cracking. | Baiyun; Tianhe; Yuexiu; Liwan; Haizhu |
| | Hilly terrain | Soft soil | High water content; highly compressible; low shear strength; prone to uneven settlement and cracking | Panyu; Huangpu |
| | Mountainous area | Lateritic red soil | High permeability; low shear strength and bearing capacity; prone to softening, collapse, and landslide. | Zencheng |

## 3. Network training

The dataset was divided into training and testing sets using a stratified random sampling strategy with a ratio of 80% to 20%, ensuring that each category keeps approximately the same proportion in both subsets. The training subset covers all the four kinds of samples with salient characteristics, while the testing subset, strictly non-overlapping with the training subset, includes various samples to evaluate the generalization capability of the three models.

Table 4 summarizes the key training settings, including data augmentation probabilities, learning rate schedule, optimizer configuration, and other relevant hyperparameters, along with their corresponding effects. The three models (Model-B, Model-C and Model-D) were trained independently using corresponding scan data. The training was conducted on a workstation equipped with dual Intel Xeon E5-2680v4 CPUs and eight NVIDIA Tesla M40 GPUs. Each model was trained for 400 epochs to guarantee convergence. Fig. 6 shows the evolution of loss during the training of the three models. The loss curves decrease steeply in the first 80 epochs and then gradually reach platform with low values. The convergence of Model-D seems poorer than Model-B and Model-C, probably because the task of distinguishing between voids and loose structures is more difficult. The duration of training was 819 minutes for Model-B, 876 minutes for Model-C, and 609 minutes for Model-D. The time consumed by the three models for inference is much less, falling in a range between 220 – 207 ms per image.

**Table 4.** Key parameters for model training

| | Parameter | Description | Effect |
|---|---|---|---|
| Data Augmentation | mosaic_prob = 1.0 | Probability of applying Mosaic augmentation. | Combines four images to improve detection performance on small objects. |
| | mixup_prob = 1.0 | Probability of applying MixUp augmentation. | Blends two images together to enhance robustness and generalization. |
| | flip_prob = 0.5 | Probability of image flipping. | Increases robustness to different road orientations. |
| | degrees = 15° | Rotation angle range. | Ensures stable recognition under various road extension directions. |
| Learning configuration | warmup_epochs = 5 | Number of epochs for linear learning rate warm-up. | Prevents instability in early training. |
| | warmup_lr = 0 | Minimum learning rate during warm-up. | |
| | min_lr_ratio = 0.05 | Minimum learning rate. | |
| | max_epoch = 400 | Maximum number of training epochs. | |
| | no_aug_epochs = 15 | Number of final epochs without augmentations such as Mosaic. | Augmentations are disabled in the last few epochs to stabilize training and improve final convergence. |
| | ema = True | Use of Exponential Moving Average (EMA) during training. | Improves robustness and stability. |
| Optimizer | weight_decay = $5\times10^{-4}$ | Optimizer weight decay. | Suppresses overfitting by penalizing large weights. |
| | momentum = 0.9 | Optimizer momentum. | Controls the influence of past gradients to smooth the optimization process. |
| | optimizer_type = AdamW | Adam with weight decay | Separates weight decay from gradient updates and applies it explicitly for better regularization and stable convergence. |

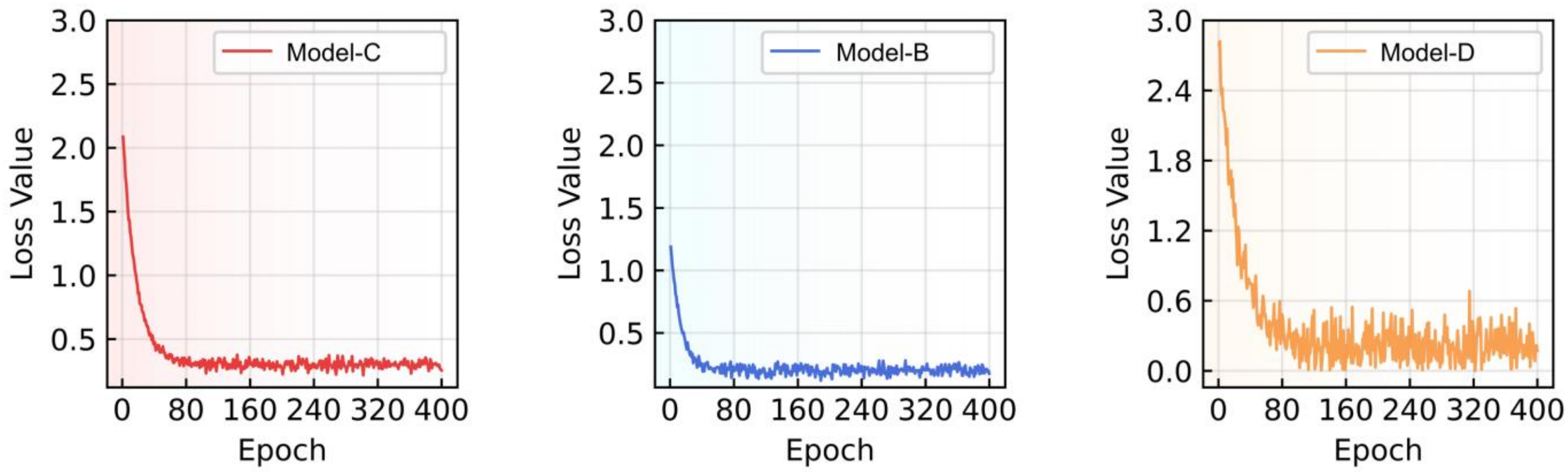

**Figure 6.** Evolution of loss during the training of Model-C, Model-B and Model-D.

# 4. Verification

## 4.1. Evaluation using testing subset

Intersection over union (IoU) is one of the most popular criteria for bounding box regression in object detection. It represents the ratio of the intersection area to the union area between the predicted bounding box $B^P$ and the ground truth $B^g$:

$$\text{IoU} = \frac{|B^g \cap B^p|}{|B^g \cup B^p|} \tag{1}$$

In this study, a predicted result is considered true if the IoU is not less than 0.5. Otherwise, it is regarded false. Afterward, precision and recall, achieved by each model individually and by the cross-verification strategy, are calculated for performance evaluation, as listed in Table 5 and Table 6.

Model-C demonstrates outstanding performance in discriminating between healthy segments and the segments containing odd objects. Its precision and recall for identifying healthy segments reach 100% and 91.8% respectively, with no segment having distress misidentified as healthy one. However, Model-C is less sensitive to manhole and distress, in particular void. The achieved precision and recall for recognition of manhole and distress vary in a range from 50.0% to 76.8%. Only about 32% of voids can be recognized by this model.

Model-B reaches precision of 97.1% and recall of 100% in discriminating manhole from distress, effectively eliminating the disturbance of manhole to RSD detection. Although Model-B can recognize most of distress, it is hard for the model to distinguish voids from loose structures, as the two types of distress look similar in B-scan.

Model-D, compared with its counterparts, achieves better differentiation between the two types of RSD. Especially, its precision and recall in identifying voids reach 85.9% and 79.4%, considerably higher than that by Model-B and Model-C.

**Table 5.** Precision achieved by the three models and by the cross-validation strategy.

| Model | Healthy parts | Manhole | Distress | Loose structure | Void |
|---|---|---|---|---|---|
| Model-C | **100%** | 50.0% | 76.8% | 60.7% | 32.4% |
| Model-B | 96.7% | **97.1%** | 90.0% | 56.7% | 67.6% |
| Model-D | 94.3% | 95.2% | **81.7%** | **59.9%** | **85.9%** |
| Cross-verification Step1 | **100%** | 50.0% | 76.8% | 60.7% | 32.4% |
| Cross-verification Step2 | | **97.1%** | 95.8% | 64.2% | 67.6% |
| Cross-verification Step3 | | | **95.9%** | **78.0%** | **88.2%** |

**Table 6.** Recall achieved by the three models and by the cross-validation strategy.

| Model | Healthy parts | Manhole | Distress | Loose structure | Void |
|---|---|---|---|---|---|
| Model-C | **91.8%** | 64.4% | 69.6% | 50.5% | 31.8% |
| Model-B | 79.1% | **100%** | 97.2% | 63.6% | 70.1% |
| Model-D | 60.0% | 98.0% | **95.8%** | **85.0%** | **79.4%** |
| Cross-verification Step1 | **91.8%** | 64.4% | 69.6% | 50.5% | 31.8% |
| Cross-verification Step2 | | **100%** | 97.2% | 63.6% | 70.1% |
| Cross-verification Step3 | | | **98.6%** | **86.0%** | **84.1%** |

The cross-verification strategy leverages the respective advantages of the three models. In Step 1, Model-C screens out 91.8% of the healthy segments in the testing subset, achieving precision and recall of 76.8% and 69.6% respectively for RSD detection. In Step 2, Model-B screens out all the manhole from the remaining samples. It is found that the precision and recall of RSD detection increase to 95.8% and 97.2%, respectively. In Step 3, application of Model-D to the remaining samples in Step 2 helps further improve the precision and recall of RSD detection to 95.9% and 98.6%, respectively. Moreover, the accuracy of identification of RSD type is also significantly enhanced. According to the recall reached in Step 3, 84.1% of the voids and 86.0% of the loose structures are correctly recognized. The proposed method outperforms most of existing approaches, especially those trained and tested using relatively large dataset from field survey (see Table 1).

### 4.2. Verification in field tests

To further verify the applicability and stability of the proposed method in practical applications, tests were conducted on new GPR survey data collected from 15 roads with total length of 69.416 km in three cities of Guangdong province (Zhuhai, Guangzhou and Yunfu). Table 7 compares the number of RSD identified by experienced inspectors and by our method. The cross-verification strategy achieves recall of 100%, without any distress missed across all the 15 roads.

**Table 7.** Number of RSD identified manually and automatically from GPR images of 15 roads.

| City | Spot | Length [km] | Number of manually identified RSD | Number of automatically identified RSD | Number of false positives |
|---|---|---|---|---|---|
| Zhuhai | Changlong Avenue | 10.400 | 7 | 34 | 27 |
| Zhuhai | Huandaodong Road | 7.000 | 4 | 4 | 0 |
| Zhuhai | Hengqindong Road | 6.000 | 0 | 6 | 6 |
| Zhuhai | Hengqinxi Road | 3.680 | 2 | 7 | 5 |
| Zhuhai | Zhongxin Avenue | 2.100 | 0 | 7 | 7 |
| Zhuhai | Jinquxi Road | 0.362 | 0 | 5 | 5 |
| Guangzhou | Daguan Road | 6.996 | 14 | 36 | 22 |
| Guangzhou | Yinglong Road | 6.400 | 11 | 27 | 15 |
| Guangzhou | Longdongdong Road | 4.800 | 5 | 12 | 7 |
| Guangzhou | Longfeng Road | 4.800 | 3 | 7 | 4 |
| Guangzhou | Yuangangheng Road | 4.200 | 16 | 36 | 20 |
| Guangzhou | Kemulang Road | 4.200 | 3 | 6 | 3 |
| Guangzhou | Nanwan Avenue | 3.980 | 3 | 4 | 1 |
| Guangzhou | Jinyuan Avenue | 3.498 | 1 | 6 | 5 |
| Yunfu | Xiangshunhuayuan Road | 1.000 | 0 | 5 | 5 |
| Total | 15 | 69.416 | 75 | 202 | 127 |

127 defects among 202 automatically detected RSDs were identified as false positives after double check by inspectors, i.e., a few healthy parts or manholes were recognized as distress. It can be found that a small number of road sections contributed disproportionately to the total number of false positives. Among all the 15 tested roads, four roads (26.7% of all roads) provide 84 false positive cases, representing 66.1% of all false positives. For example, the automatically detected RSD in Changlong Avenue (Zhuhai) is as much as 4.9 times the manual results. It is noteworthy that the automatic detection and the manual detection agree completely in nearby Huandaodong Road (Zhuhai), indicating the probably incipient risk in Changlong Avenue.

This outcome is consistent with the design philosophy of our method, which prioritizes sensitivity to minimize the risk of overlooking potential RSD. The conservative configuration inevitably results in higher proportion of false positives. Nevertheless, the overestimated RSD brings limited impact, as the manual review of these cases only takes a few additional minutes. Moreover, these road sections with more plausible RSD may require frequent and careful inspection. The abnormally high proportion of false positives could serve as a useful indicator of potential high-risk sections, helping to distribute monitoring and maintenance resources efficiently.

Table 8 compares the time consumed by manual processing and automatic processing. Manual processing took a total of 1128 minutes (about 19 hours, namely 2.4 working days). In contrast, reviewing the results of automatic processing takes only 101 minutes (less than 2 hours), which demonstrates a massive saving of labor for approximately 90%. It is noteworthy that the automatic processing seems even less efficient than manual processing, taking 1560 minutes to check all the GPR images of 15 roads. The main reason is the tests were run on a workstation (costs around 5,000 USD) with much lower computational performance in comparison with up-to-date workstations (cost five times higher). Moreover, the cost-effectiveness of hardware allows the deployment of multiple workstations, which can greatly accelerate the processing of data through parallelism.

**Table 8** Time consumption of manual and automatic RSD detection, as well as the combination.

| City | Spot | Manual [minute] | Automatic [minute] | Manual review of automatic recognition [minute] |
|---|---|---|---|---|
| Zhuhai | Changlong Avenue | 126 | 138 | 17 |
| Zhuhai | Huandaodong Road | 90 | 66 | 2 |
| Zhuhai | Hengqindong Road | 90 | 102 | 3 |
| Zhuhai | Hengqinxi Road | 48 | 84 | 3.5 |
| Zhuhai | Zhongxin Avenue | 36 | 84 | 3.5 |
| Zhuhai | Jinquxi Road | 36 | 120 | 2.5 |
| Guangzhou | Daguan Road | 120 | 114 | 18 |
| Guangzhou | Yinglong Road | 78 | 96 | 13.5 |
| Guangzhou | Longdongdong Road | 42 | 78 | 6 |
| Guangzhou | Longfeng Road | 42 | 84 | 3.5 |
| Guangzhou | Yuangangheng Road | 120 | 168 | 18 |
| Guangzhou | Kemulang Road | 120 | 180 | 3 |
| Guangzhou | Nanwan Avenue | 90 | 84 | 2 |
| Guangzhou | Jinyuan Avenue | 48 | 72 | 3 |
| Yunfu | Xiangshunhuayuan Road | 42 | 90 | 2.5 |
| Total | 15 | 1128 | 1560 | 101 |

## 5. Conclusion and outlook

This study demonstrates an efficient and reliable approach to automatically detect road subsurface distress from 3D GPR data.

Based on the finding that the YOLO-based model demonstrates varying sensitivity to specific RSD in B-scan, C-scan, and D-scan, respectively. A novel cross-verification strategy is proposed to leverage the respective strength of the model working on different scans, in which three models trained independently with the three scans were combined sequentially in automatic RSD recognition. The cross-verification strategy reached recall and precision of 98.6% and 95.9% respectively in RSD detection on the dataset from the field surveys of 105 typical urban road sections (1250 km) in two metropolises of China (Chengdu and Guangzhou). In field tests conducted on 15 road sections (69.4 km) in three Chinese cities (Zhuhai, Guangzhou and Yunfu), it achieved recall of 100% without any RSD overlooked. More importantly, experiments show that a new work procedure combining automatic processing of GPR data followed by manual review can reduce the labor of inspection by about 90% without any loss of reliability of RSD recognition. It well meets the requirement in large-scale and long-term maintenance of current urban roads.

The proposed cross-verification strategy is designed to be independent of object detection model, which enables the flexible combination with a wide range of deep learning models and facilitates its use as a benchmark framework for evaluating RSD detection algorithms on larger and more diverse GPR datasets.

It is noteworthy that the current configuration prioritizes recall to ensure exhaustive detection of potential RSD, the confidence threshold could be adjusted to accommodate different service conditions and inspection objectives. This flexibility allows the method to operate effectively across various field scenarios with different trade-offs between precision and recall.

## Acknowledgments

The study was financially supported by the National Natural Science Foundation of China (Grant Nos. 12232017, 12472179 and 12502215), Natural Science Foundation of Guangdong Province (Grant No. 2024A1515011076), Guangdong Provincial Key Laboratory of Intelligent Disaster Prevention and Emergency Technologies for Urban Lifeline Engineering (Grant No. 2022B1212010016), Guangzhou Science and Technology Planning Project (Grant No. 2024A04J4348), and the Postdoctoral Research Foundation of China (2021M700886).

## References

[1] G. Yue, Y. Du, C. Liu, S. Guo, Y. Li, Q. Gao, Road subsurface distress recognition method using multiattribute feature fusion with ground penetrating radar, International Journal of Pavement Engineering 24 (2023) 2037591. https://doi.org/10.1080/10298436.2022.2037591.

[2] Z. Zhu, G. Zang, G. Jin, W. Cai, Z. Zhang, Quantitative Evaluation Method for Asphalt Pavement Structure Integrity Based on Ground Penetrating Radar, in: CICTP 2020, American Society of Civil Engineers, Xi'an, China (Conference Cancelled), 2020: pp. 1212–1221. https://doi.org/10.1061/9780784483053.102.

[3] Z. Tong, J. Gao, D. Yuan, Advances of deep learning applications in ground-penetrating radar: A survey, Construction and Building Materials 258 (2020) 120371. https://doi.org/10.1016/j.conbuildmat.2020.120371.

[4] C. Liu, Y. Du, G. Yue, Y. Li, D. Wu, F. Li, Advances in automatic identification of road subsurface distress using ground penetrating radar: State of the art and future trends, Automation in Construction 158 (2024) 105185. https://doi.org/10.1016/j.autcon.2023.105185.

[5] X. Xu, S. Peng, Y. Xia, W. Ji, The development of a multi-channel GPR system for roadbed damage detection, Microelectronics Journal 45 (2014) 1542–1555. https://doi.org/10.1016/j.mejo.2014.09.004.

[6] A. Srivastav, P. Nguyen, M. McConnell, K.A. Loparo, S. Mandal, A Highly Digital Multiantenna Ground-Penetrating Radar (GPR) System, IEEE Trans. Instrum. Meas. 69 (2020) 7422–7436. https://doi.org/10.1109/TIM.2020.2984415.

[7] M. Rasol, J.C. Pais, V. Pérez-Gracia, M. Solla, F.M. Fernandes, S. Fontul, D. Ayala-Cabrera, F. Schmidt, H. Assadollahi, GPR monitoring for road transport infrastructure: A systematic review and machine learning insights, Construction and Building Materials 324 (2022) 126686. https://doi.org/10.1016/j.conbuildmat.2022.126686.

[8] Z. Du, J. Yuan, F. Xiao, C. Hettiarachchi, Application of image technology on pavement distress detection: A review, Measurement 184 (2021) 109900. https://doi.org/10.1016/j.measurement.2021.109900.

[9] X.L. Travassos, S.L. Avila, N. Ida, Artificial Neural Networks and Machine Learning techniques applied to Ground Penetrating Radar: A review, ACI 17 (2021) 296–308. https://doi.org/10.1016/j.aci.2018.10.001.

[10] X. Xiong, A. Meng, J. Lu, Y. Tan, B. Chen, J. Tang, C. Zhang, S. Xiao, J. Hu, Automatic detection and location of pavement internal distresses from ground penetrating radar images based on deep learning, Construction and Building Materials 411 (2024) 134483. https://doi.org/10.1016/j.conbuildmat.2023.134483.

[11] W. Al-Nuaimy, Y. Huang, M. Nakhkash, M.T.C. Fang, V.T. Nguyen, A. Eriksen, Automatic detection of buried utilities and solid objects with GPR using neural networks and pattern recognition, Journal of Applied Geophysics 43 (2000) 157–165. https://doi.org/10.1016/S0926-9851(99)00055-5.

[12] C. Maas, J. Schmalzl, Using pattern recognition to automatically localize reflection hyperbolas in data from ground penetrating radar, Computers & Geosciences 58 (2013) 116–125. https://doi.org/10.1016/j.cageo.2013.04.012.

[13] Wai-Lok Lai, Chi-Sun Poon, GPR data analysis in time-frequency domain, in: 2012 14th International Conference on Ground Penetrating Radar (GPR), IEEE, Shanghai, 2012: pp. 362–366. https://doi.org/10.1109/ICGPR.2012.6254891.

[14] S.S. Todkar, C. Le Bastard, V. Baltazart, A. Ihamouten, X. Dérobert, Performance assessment of SVM-based classification techniques for the detection of artificial debondings within pavement structures from stepped-frequency A-scan radar data, NDT & E International 107 (2019) 102128. https://doi.org/10.1016/j.ndteint.2019.102128.

[15] H. Ali, A.F.A. Zaidi, C.E. Meng, M.R. Ahmad, T.S.T. Amran, S.N.A.M. Kanafiah, C.Y. Fook, S.A.A. Shukor, M. Elshaikh, Shape classification of ground penetrating radar using discrete wavelet

transform and principle component analysis, IOP Conf. Ser.: Mater. Sci. Eng. 705 (2019) 012046. https://doi.org/10.1088/1757-899X/705/1/012046.
[16] X. Zhang, J. Bolton, P. Gader, A New Learning Method for Continuous Hidden Markov Models for Subsurface Landmine Detection in Ground Penetrating Radar, IEEE J. Sel. Top. Appl. Earth Observations Remote Sensing 7 (2014) 813–819. https://doi.org/10.1109/JSTARS.2014.2305981.
[17] H. Harkat, A.E. Ruano, M.G. Ruano, S.D. Bennani, GPR target detection using a neural network classifier designed by a multi-objective genetic algorithm, Applied Soft Computing 79 (2019) 310–325. https://doi.org/10.1016/j.asoc.2019.03.030.
[18] Z. Xiang, A. Rashidi, G. (Gaby) Ou, An Improved Convolutional Neural Network System for Automatically Detecting Rebar in GPR Data, in: Computing in Civil Engineering 2019, American Society of Civil Engineers, Atlanta, Georgia, 2019: pp. 422–429. https://doi.org/10.1061/9780784482438.054.
[19] U. Özkaya, Ş. Öztürk, F. Melgani, L. Seyfi, Residual CNN + Bi-LSTM model to analyze GPR B scan images, Automation in Construction 123 (2021) 103525. https://doi.org/10.1016/j.autcon.2020.103525.
[20] M.M. Rosso, G. Marasco, S. Aiello, A. Aloisio, B. Chiaia, G.C. Marano, Convolutional networks and transformers for intelligent road tunnel investigations, Computers & Structures 275 (2023) 106918. https://doi.org/10.1016/j.compstruc.2022.106918.
[21] W. Lei, F. Hou, J. Xi, Q. Tan, M. Xu, X. Jiang, G. Liu, Q. Gu, Automatic hyperbola detection and fitting in GPR B-scan image, Automation in Construction 106 (2019) 102839. https://doi.org/10.1016/j.autcon.2019.102839.
[22] J. Redmon, S. Divvala, R. Girshick, A. Farhadi, You Only Look Once: Unified, Real-Time Object Detection, in: 2016 IEEE Conference on Computer Vision and Pattern Recognition (CVPR), IEEE, Las Vegas, NV, USA, 2016: pp. 779–788. https://doi.org/10.1109/CVPR.2016.91.
[23] Y. Yue, H. Liu, C. Lin, X. Meng, C. Liu, X. Zhang, J. Cui, Y. Du, Automatic recognition of defects behind railway tunnel linings in GPR images using transfer learning, Measurement 224 (2024) 113903. https://doi.org/10.1016/j.measurement.2023.113903.
[24] Z. Qiu, Z. Zhao, S. Chen, J. Zeng, Y. Huang, B. Xiang, Application of an Improved YOLOv5 Algorithm in Real-Time Detection of Foreign Objects by Ground Penetrating Radar, Remote Sensing 14 (2022) 1895. https://doi.org/10.3390/rs14081895.
[25] J. Zhang, X. Yang, W. Li, S. Zhang, Y. Jia, Automatic detection of moisture damages in asphalt pavements from GPR data with deep CNN and IRS method, Automation in Construction 113 (2020) 103119. https://doi.org/10.1016/j.autcon.2020.103119.
[26] Z. Liu, X. Gu, H. Yang, L. Wang, Y. Chen, D. Wang, Novel YOLOv3 Model With Structure and Hyperparameter Optimization for Detection of Pavement Concealed Cracks in GPR Images, IEEE Trans. Intell. Transport. Syst. 23 (2022) 22258–22268. https://doi.org/10.1109/TITS.2022.3174626.
[27] Y. Li, Z. Zhao, Y. Luo, Z. Qiu, Real-Time Pattern-Recognition of GPR Images with YOLO v3 Implemented by Tensorflow, Sensors 20 (2020) 6476. https://doi.org/10.3390/s20226476.
[28] Y. Li, C. Liu, G. Yue, Q. Gao, Y. Du, Deep learning-based pavement subsurface distress detection via ground penetrating radar data, Automation in Construction 142 (2022) 104516. https://doi.org/10.1016/j.autcon.2022.104516.
[29] H. Liu, Y. Yue, C. Liu, B.F. Spencer, J. Cui, Automatic recognition and localization of underground pipelines in GPR B-scans using a deep learning model, Tunnelling and Underground Space Technology 134 (2023) 104861. https://doi.org/10.1016/j.tust.2022.104861.
[30] R. Liu, Y. Li, P. Yin, H. Sun, Z. Bao, X. Yang, Layered Media Inversion Network Applied in Ground Penetrating Radar, in: 2021 CIE International Conference on Radar (Radar), IEEE, Haikou, Hainan, China, 2021: pp. 2196–2199. https://doi.org/10.1109/Radar53847.2021.10028533.
[31] N. Li, R. Wu, H. Li, H. Wang, Z. Gui, D. Song, MV-GPRNet: Multi-View Subsurface Defect Detection Network for Airport Runway Inspection Based on GPR, Remote Sensing 14 (2022) 4472. https://doi.org/10.3390/rs14184472.

[32] F. Li, F. Yang, X. Qiao, Z. Hu, X. Wu, H. Xing, 3D ground penetrating radar road underground target identification algorithm using time-frequency statistical features of data, NDT & E International 137 (2023) 102860. https://doi.org/10.1016/j.ndteint.2023.102860.
[33] S. Khudoyarov, N. Kim, J.-J. Lee, Three-dimensional convolutional neural network–based underground object classification using three-dimensional ground penetrating radar data, Structural Health Monitoring 19 (2020) 1884–1893. https://doi.org/10.1177/1475921720902700.
[34] T. Yamaguchi, T. Mizutani, T. Nagayama, Mapping Subsurface Utility Pipes by 3-D Convolutional Neural Network and Kirchhoff Migration Using GPR Images, IEEE Trans. Geosci. Remote Sensing 59 (2021) 6525–6536. https://doi.org/10.1109/TGRS.2020.3030079.
[35] T. Yamaguchi, T. Mizutani, K. Meguro, T. Hirano, Detecting Subsurface Voids From GPR Images by 3-D Convolutional Neural Network Using 2-D Finite Difference Time Domain Method, IEEE J. Sel. Top. Appl. Earth Observations Remote Sensing 15 (2022) 3061–3073. https://doi.org/10.1109/JSTARS.2022.3165660.
[36] M.-S. Kang, N. Kim, S.B. Im, J.-J. Lee, Y.-K. An, 3D GPR Image-based UcNet for Enhancing Underground Cavity Detectability, Remote Sensing 11 (2019) 2545. https://doi.org/10.3390/rs11212545.
[37] N. Kim, S. Kim, Y.-K. An, J.-J. Lee, A novel 3D GPR image arrangement for deep learning-based underground object classification, International Journal of Pavement Engineering 22 (2021) 740–751. https://doi.org/10.1080/10298436.2019.1645846.
[38] N. Kim, S. Kim, Y.-K. An, J.-J. Lee, Triplanar Imaging of 3-D GPR Data for Deep-Learning-Based Underground Object Detection, IEEE J. Sel. Top. Appl. Earth Observations Remote Sensing 12 (2019) 4446–4456. https://doi.org/10.1109/JSTARS.2019.2953505.
[39] Z. Liu, X. Gu, J. Chen, D. Wang, Y. Chen, L. Wang, Automatic recognition of pavement cracks from combined GPR B-scan and C-scan images using multiscale feature fusion deep neural networks, Automation in Construction 146 (2023) 104698. https://doi.org/10.1016/j.autcon.2022.104698.
[40] J. Yang, K. Ruan, J. Gao, S. Yang, L. Zhang, Pavement Distress Detection Using Three-Dimension Ground Penetrating Radar and Deep Learning, Applied Sciences 12 (2022) 5738. https://doi.org/10.3390/app12115738.
[41] H. Lv, Y. Zhang, J. Dai, H. Wu, J. Wang, D. Wang, Intelligent Recognition of GPR Road Hidden Defect Images Based on Feature Fusion and Attention Mechanism, IEEE Trans. Geosci. Remote Sensing 63 (2025) 1–17. https://doi.org/10.1109/tgrs.2025.3575293.
[42] Z. Ge, S. Liu, F. Wang, Z. Li, J. Sun, YOLOX: Exceeding YOLO Series in 2021, (2021). https://doi.org/10.48550/ARXIV.2107.08430.
[43] G. Jocher, Ayush Chaurasia, A. Stoken, J. Borovec, NanoCode012, Yonghye Kwon, Kalen Michael, TaoXie, Jiacong Fang, Imyhxy, Lorna, Zeng Yifu, C. Wong, Abhiram V, D. Montes, Zhiqiang Wang, C. Fati, Jebastin Nadar, Laughing, UnglvKitDe, V. Sonck, Tkianai, YxNONG, P. Skalski, A. Hogan, Dhruv Nair, M. Strobel, M. Jain, ultralytics/yolov5: v7.0 - YOLOv5 SOTA Realtime Instance Segmentation, (2022). https://doi.org/10.5281/ZENODO.3908559.
[44] K. He, X. Zhang, S. Ren, J. Sun, Spatial Pyramid Pooling in Deep Convolutional Networks for Visual Recognition, IEEE Trans. Pattern Anal. Mach. Intell. 37 (2015) 1904–1916. https://doi.org/10.1109/TPAMI.2015.2389824.
[45] C.-Y. Wang, H.-Y. Mark Liao, Y.-H. Wu, P.-Y. Chen, J.-W. Hsieh, I.-H. Yeh, CSPNet: A New Backbone that can Enhance Learning Capability of CNN, in: 2020 IEEE/CVF Conference on Computer Vision and Pattern Recognition Workshops (CVPRW), IEEE, Seattle, WA, USA, 2020: pp. 1571–1580. https://doi.org/10.1109/CVPRW50498.2020.00203.
[46] K. He, X. Zhang, S. Ren, J. Sun, Deep Residual Learning for Image Recognition, in: 2016 IEEE Conference on Computer Vision and Pattern Recognition (CVPR), IEEE, Las Vegas, NV, USA, 2016: pp. 770–778. https://doi.org/10.1109/CVPR.2016.90.

[47] P.F. Felzenszwalb, R.B. Girshick, D. McAllester, D. Ramanan, Object Detection with Discriminatively Trained Part-Based Models, IEEE Trans. Pattern Anal. Mach. Intell. 32 (2010) 1627–1645. https://doi.org/10.1109/TPAMI.2009.167.